\documentclass[sigconf]{acmart}

\AtBeginDocument{%
  }

\copyrightyear{2022}
\acmYear{2022}
\setcopyright{acmcopyright}\acmConference[KDD '22]{Proceedings of the 28th ACM SIGKDD Conference on Knowledge Discovery and Data Mining}{August 14--18, 2022}{Washington, DC, USA}
\acmBooktitle{Proceedings of the 28th ACM SIGKDD Conference on Knowledge Discovery and Data Mining (KDD '22), August 14--18, 2022, Washington, DC, USA}
\acmPrice{15.00}
\acmDOI{10.1145/3534678.3539258}
\acmISBN{978-1-4503-9385-0/22/08}

\usepackage{amsmath}
\usepackage{array}

\usepackage{graphicx}
\usepackage{algorithm}
\usepackage{algorithmic}
\usepackage{booktabs}
\usepackage{multirow}
\usepackage{amsfonts}

\usepackage{nicefrac}

\newcommand{\R}{\mathbb{R}}

\makeatletter
\newcommand{\ssymbol}[1]{}
\makeatother

\begin{document}

\title{Graph-in-Graph Network for Automatic Gene Ontology Description Generation}

\author{Fenglin Liu}
\affiliation{%
  \institution{School of ECE, \\Peking University}
  \country{China}}
\email{fenglinliu98@pku.edu.cn}

\author{Bang Yang}
\affiliation{%
  \institution{School of ECE, \\Peking University}
  \country{China}}
\email{yangbang@pku.edu.cn}

\author{Chenyu You}
\affiliation{%
  \institution{Department of Electrical Engineering,\\ Yale University}
  \country{USA}}
\email{chenyu.you@yale.edu}

\author{Xian Wu}
\affiliation{\institution{Jarvis Lab, \\ Tencent}
  \country{China}}
\email{kevinxwu@tencent.com}

\author{Shen Ge}
\affiliation{\institution{Jarvis Lab, \\ Tencent}
  \country{China}}
\email{shenge@tencent.com}

\author{Adelaide Woicik}
\affiliation{%
  \institution{Paul G. Allen School of Computer Science, University of Washington}
  \country{USA}}
\email{addiewc@cs.washington.edu}

\author{Sheng Wang}
\authornote{Corresponding author.}
\affiliation{%
  \institution{Paul G. Allen School of Computer Science, University of Washington}
  \country{USA}}
\email{swang@cs.washington.edu}

\renewcommand{\shortauthors}{Fenglin Liu et al.}
\begin{abstract}
Gene Ontology (GO) is the primary gene function knowledge base that enables computational tasks in biomedicine. The basic element of GO is a term, which includes a set of genes with the same function. Existing research efforts of GO mainly focus on predicting gene term associations. Other tasks, such as generating descriptions of new terms, are rarely pursued. In this paper, we propose a novel task: GO term description generation. This task aims to automatically generate a sentence that describes the function of a GO term belonging to one of the three categories, i.e., molecular function, biological process, and cellular component. To address this task, we propose a Graph-in-Graph network that can efficiently leverage the structural information of GO. The proposed network introduces a two-layer graph: the first layer is a graph of GO terms where each node is also a graph (gene graph). Such a Graph-in-Graph network can derive the biological functions of GO terms and generate proper descriptions. To validate the effectiveness of the proposed network, we build three large-scale benchmark datasets. By incorporating the proposed Graph-in-Graph network, the performances of seven different sequence-to-sequence models can be substantially boosted across all evaluation metrics, with up to 34.7\%, 14.5\%, and 39.1\% relative improvements in BLEU, ROUGE-L, and METEOR, respectively.
\end{abstract}



\maketitle

\section{Introduction}

\begin{figure}[t]
\centering
\includegraphics[width=1\linewidth]{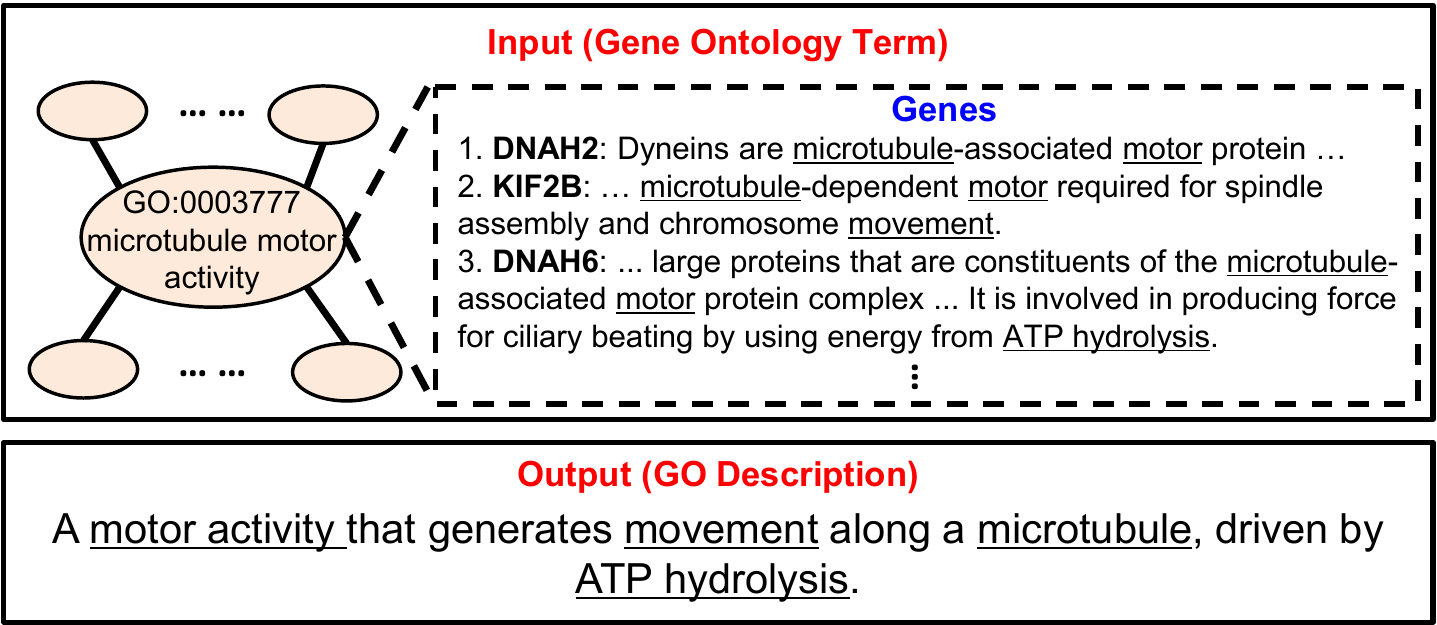}
\caption{Examples of the GO term “microtubule motor activity” and its associated GO description. The term contains a set of genes DNAH2, KIF2B, DNAH6..., which are annotated with the gene text, and the GO description provides the biological function of combined genes. The GO structure around the term of interest is organized hierarchically, with other terms shown in orange ovals.}
\label{fig:introduction}
\end{figure}

Gene Ontology (GO), which includes tens of thousands of biological functions, is crucial to biomedical research \cite{ashburner2000gene, dutkowski2013gene} and advances many downstream applications, such as protein function prediction \cite{wang2015exploiting}, disease mechanism analysis \cite{davis2016generating} and drug discovery \cite{mutowo2016drug}. Biological functions, which we refer to as Gene Ontology \textit{terms}, are grouped and organized hierarchically according to three categories: molecular function, biological process and cellular component \cite{GO2015GO,GO2017GO}, where each node is a term and each edge represents an `is a' relationship between two terms (Figure~\ref{fig:introduction}). Each term is further associated with a set of genes that have this function and a curated text description of this function.

Although GO has been used extensively, a key challenge for using GO is the updating of GO with new biomedical discoveries, which are often part of newly-generated biomedical literature \cite{tomczak2018interpretation, xu2022protranslator}. Currently, GO updates are performed manually by domain experts through digesting scientific literature, annotating associated genes, identifying similar terms and writing descriptions. As this multi-step process is tedious and time-consuming, even for domain experts, expert updates can hardly scale to the size of GO.
However, to the best of our knowledge, there are no existing works attempted to generate the description of a GO term, which is one of the most time-consuming steps for updating GO. 
More importantly, the inconsistency in the descriptions generated by different experts may further cause bias in the GO description and hinder the process of GO analysis. Therefore, there is a pressing need to automatically generate GO term descriptions in order to assist biologists in discovering the gene function of new terms \cite{Elham2017Survey,liu2021graphine,yangpathway2text}.

To this end, we propose the novel task of automatic GO term description generation, which aims to generate a textual description for an input term based on the associated genes. After examining the GO data, we find two types of information that can be leveraged to model the description: 1) \textit{Term graph structure:} terms in GO are organized hierarchically, which enables us to capture their biological relationships by representing terms as a directed acyclic graph; 2) \textit{Overlapping phrases}: many phrases are reused in both gene text and the term description for a given GO term. We illustrated one such case in Figure \ref{fig:introduction} where the overlapping phrases are underlined. The overlapping phrases further enables us to construct a \textit{gene graph} using these phrases. At last, to model the term graph and gene graph simultaneously, we propose a graph-in-graph network approach. As shown in Figure~\ref{fig:framework}, the outside layer graph is the term graph. Nodes in the term graph also form an inside layer graph, where the nodes are the term itself, genes belonging to this term, and individual words in the gene text. Such a graph-in-graph network can capture both the biological relations between terms and the gene relations within each term, providing a holistic view of the GO data. By encoding and decoding information in the graph-in-graph network, we can generate the description for a novel GO term as soon as it is inserted into the term graph.  
It can assist biologists in GO construction and analysis. For domain experts, given a new GO term with its associated genes, our model can automatically generate a textual description to describe its biological function, and the experts only need to make revision rather than writing a new description from scratch.
It can not only relieve the experts from heavy workload but also alert them to some important biological functions to avoid mislabeling and missed labeling (avoid errors and omissions).

To prove the effectiveness of our graph-in-graph network, we build three large-scale benchmark datasets, i.e., Molecular Function, Biological Process and Cellular Component. In our experiments, we incorporate the graph-in-graph network into seven different sequence-to-sequence models: 1) RNN-based model, 2) attention-based model, 3) hierarchical RNN-based model, 4) copy mechanism based model, 5) convolutional-neural-network- (CNN-) based model, 6) graph-to-sequence model, and 7) fully-attentive model, i.e., Transformer \cite{Vaswani2017Transformer}. The experiments show that our approach can substantially boost the performance of baselines across all evaluation metrics.

Overall, the main contributions of this paper are:
\begin{itemize}

    \item We make the first attempt to automatically describe the biological function of GO terms, which can assist biologists in GO construction and analysis. To address the task, we build three large-scale benchmark datasets and propose a novel Graph-in-Graph network.

    \item Our approach introduces a gene graph to model the semantic similarity between genes within a term and a term graph to model the structural similarity between terms. Two graphs are coupled together to generate the GO term description.

    \item We prove the effectiveness and the generalization capabilities of our approach on three datasets. After including our Graph-in-Graph network, performances of the baseline models improve significantly on all metrics, 
    with up to 34.7\%, 14.5\%, and 39.1\% relative improvements in BLEU, ROUGE-L, and METEOR, respectively. The analysis further highlights the transferability of our approach.
\end{itemize}

\section{Related Works}
In this section, we describe the related works: 1) Gene Ontology Construction; 2) Sequence-to-Sequence Learning.

\subsection{Gene Ontology Construction}
Recently, several computational approaches have been proposed to automate Gene Ontology construction. 
\citet{Mazandu2017Gene} builds and organizes semantically-related terms; \citet{koopmans2019syngo} helps identify the relationship between newly-discovered terms and existing terms; \citet{Zhang2020GO} proposes to generate the name of new discovered terms. 
Specifically, NeXO \cite{dutkowski2013gene} is proposed to reconstruct the graph structure of gene ontology by clustering genes based on their topological features encoded in the molecular networks. CliXO  \cite{Kramer2014Inferring} further incorporated multi-omics and semantic information into the clustering procedure and obtained an improved performance in GO construction. A few other approaches have proposed to predict term associations on the Gene Ontology by utilizing its graph structure and modeling it as a link prediction problem \cite{Gligorijevic2014Integration,li2016integrating,Peng2016Extending}. Despite the encouraging performance of these methods on a variety of tasks, none of the existing methods has attempted to generate the description of a GO term, which is arguably the most time-consuming step in updating GO. Consequently, gene functions of new terms are still manually annotated by the experts, which is very time-consuming, tedious and inefficient.
\citet{Zhang2020GO} introduced a graph-based approach to automatically annotate the GO terms given their associated set of genes. However, they focus on automatic generation of the term names, e.g., ''microtubule motor activity`` in Figure~\ref{fig:introduction}, which are usually very short (five words on average). Therefore, term name generation cannot fully represent the rich functional semantic of GO terms.
Rather than generating term names, we propose to generate the descriptions of gene function, which are usually $20\sim30$ words in length (see Table~\ref{tab:statistics}), to assist biologists in discovering new GO terms and GO construction and better support research in biomedicine and biology.

\begin{figure*}[t]

\centering
\includegraphics[width=1\linewidth]{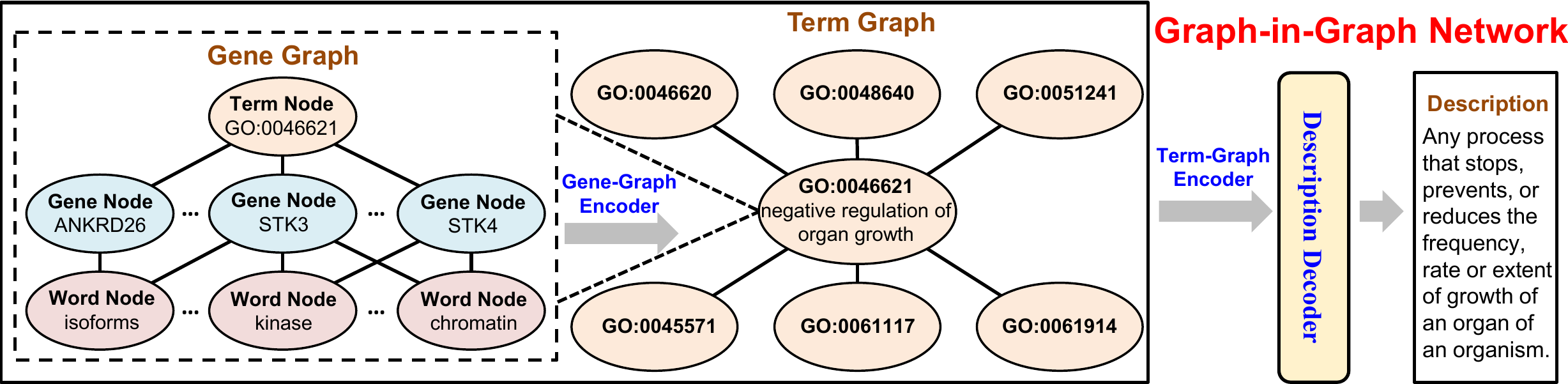}
\caption{Illustration of our Graph-in-Graph network. Specifically, we first construct a gene graph for each term to model the relationships between genes that share a specific term and further construct a tree-like term graph to model the relationships between the GO term nodes, which includes both the parent nodes (PN) and child nodes (CN). Thus, the Graph-in-Graph stands for Gene Graph-in-Term Graph. Then, our approach introduces a gene-graph encoder and a term-graph encoder to capture the structural information of Gene Ontology (GO).}
\label{fig:framework}
\end{figure*}

\subsection{Sequence-to-Sequence Learning}
In recent years, many deep neural systems have been proposed for sequence-to-sequence learning.
The common approaches \cite{Vaswani2017Transformer,bahdanau2014neural} use the encoder-decoder framework, which is usually based on the RNN or CNN model \cite{Hochreiter1997LSTM,Gehring2017ConvS2S}, to map a source sequence to a target sequence, such as in machine translation \cite{bahdanau2014neural} and paraphrasing \cite{Liu2019Paraphrase}.
The encoder network computes intermediate representations for the source sequence and the decoder network defines a probability distribution over target sentences given the intermediate representation. 
To enable a more efficient use of the source sequences, a copy mechanism \cite{Gu2016CopyNet} and a series of attention methods \cite{Luong2015Seq2Seq_Attention,Vaswani2017Transformer} have been proposed to directly provide the decoder with the source information.
In particular, the recent advent of fully-attentive models, e.g., Transformer \cite{Vaswani2017Transformer,Devlin2019BERT}, in which no recurrence is required, have been proposed and successfully applied to multiple tasks.
For the network structure, although graph-to-sequence models \cite{Kedziorski2019Text,ozturk2020exploring,Cai2020Graph} have been developed, 
in our work, to exploit the information of overlapping phrases, given the input \textit{plain text} of GO, we construct a gene graph using these overlapping phrases, while most existing graph-to-sequence tasks leverage existing explicit structures to construct the graph.

\section{Approach}
\label{sec:approach}
We first define the automatic Gene Ontology (GO) description problem. Next, we describe the proposed Graph-in-Graph network in detail.

\subsection{GO Description Problem Definition}
\label{sec:definition}
We use the pair ($T$, $S$) to denote a GO term and its corresponding textual description. Let $T=\{g_1,g_2,\dots,g_{N_\text{G}}\}$ denote the $N_\text{G}$ genes belonging to this term, where $g_i \in \R^{d}$ represents the embedding of the $i^{th}$ gene. As shown in Figure \ref{fig:introduction}, each gene has a corresponding description, i.e., gene text. To produce $g_i$, we embed the gene text with BiLSTM \cite{Hochreiter1997LSTM}. Let $S=\{y_1,y_2,\dots,y_{N_\text{S}}\}$ denote the target description of the GO term which contains $N_\text{S}$ words. The goal of our proposed GO description task is to generate $S$ given $T$.

Since both the input $T$ and the output $S$ are sequences, we can adopt the encoder-decoder framework, which is widely-used in sequence-to-sequence tasks (e.g., neural machine translation \cite{bahdanau2014neural} and paraphrasing \cite{Liu2019Paraphrase}), to perform the GO description generation task. In particular, the encoder-decoder framework includes a term encoder and a description decoder, which is formulated as:
\begin{equation}
\label{eq:definition}
\text{Term Encoder}  : T \to I ;
\quad
\text{Description Decoder} : I \to S .
\end{equation}
The term encoder, e.g., LSTM \cite{Hochreiter1997LSTM} or Transformer \cite{Vaswani2017Transformer}, aims to compute the intermediate representation $I$ from the input $T$. Then $I$ is fed into the decoder network, e.g., LSTM \cite{Hochreiter1997LSTM} or Transformer \cite{Vaswani2017Transformer}, to generate the description. The decoder defines the probability distribution $p_{\theta}\left(y_{t} \mid y_{1: t-1}, I\right)$ over target sentences $S$ given the intermediate representation $I$. Finally, given a target ground-truth sequence $\{y_1^*, ..., y_{N_\text{S}}^*\}$ and the GO description model with parameters $\theta$, the training objective is to minimize the following widely-used cross-entropy loss:
\begin{align}
\label{eq:loss}
L_{\text{XE}}(\theta)=-\sum_{t=1}^{N_\text{S}} \log \left(p_{\theta}\left(y_{t}^{*} \mid y_{1: t-1}^{*}\right)\right).
\end{align}

\subsection{Graph-in-Graph Network}
To model the relationship between terms and the relationships within a term, we propose a graph-in-graph network. As shown in Figure~\ref{fig:framework}, the outer-layer graph is the term graph $\mathcal{G}_\text{term}  = (V', E')$ which models the biological relationships between terms. We represent each node in $V'$ with a gene graph $\mathcal{G}_\text{gene} = (V, E)$, which models the relationships between genes within this term. As shown in the left plot of Figure~\ref{fig:framework}, the set of nodes $V$ in the gene graph includes the term itself, the genes belonging to this term and the words in the gene text.
To capture the nested structural information, the Graph-in-Graph network introduces two encoders, i.e., the Gene Graph Encoder and the Term Graph Encoder.
Therefore, the baseline encoder-decoder model equipped with the graph-in-graph network can be formulated as:
\begin{align}
\label{eq:formulation}
\text{Graph Construction} &: T \to \{\mathcal{G}_\text{gene}, \mathcal{G}_\text{term}\}  \nonumber \\
\text{Graph-in-Graph Network}  &:  \{\mathcal{G}_\text{gene}, \mathcal{G}_\text{term}\} \to I \nonumber   \\
\text{Description Decoder} &: I \to S . 
\end{align}
Next, we formalize our graph-in-graph construction and encoding.

\subsubsection{Gene Graph}
The gene graph is defined as $\mathcal{G}_\text{gene} = (V, E)$, where $V = \{v_i\}_{i=1:N_\text{gene}} \in \R^{N_\text{gene} \times d}$ is a set of nodes and $E = \{e_{i,j}\}_{i,j=1:N_\text{gene}}$ is a set of edges. Given a term $T=\{g_1,g_2,\dots,g_{N_\text{G}}\}$, which includes $N_G$ genes, we use the term itself, associated genes and each word in the gene text as nodes. Given $N_W$ words in the gene text for all associated genes, the number of nodes in $\mathcal{G}_\text{gene}$ is $N_\text{gene} = 1+N_G+N_W$. We include two types of edges in $E$: we first connect the term with its related genes and then connect the genes with the words extracted from their descriptions. For each node in $V$, we represent it with the embedding $v_i \in \R^{d}$. In this gene graph, the node embedding of the term node, gene nodes and word nodes are initialized with the term name embedding, the gene embeddings ($\{g_i\}$) and the word embeddings, respectively. The edge weights of all term-gene are set to 1, while the gene-word edges are set the degree-normalized co-occurrence of gene node and word node computed from the current gene text.

The gene graph is then encoded using a gene-graph encoder, defined as
\begin{equation}
\hat{v}_{i}={v}_{i}+\text{ReLU}\left(\sum\nolimits_{j=1}^{N_\text{gene}} {e}_{i, j} {W}_{v} {v}_{j}\right) ,
\end{equation}
where $\text{ReLU}(\cdot)$ represents the ReLU activation function and $W_v \in \R^{d \times d}$ is a learnable matrix. This enables us to obtain the set of encoded gene node embeddings:
\begin{align}
\label{eqn:gene_graph}
\hat{V} &= \text{Gene-Graph-Encoder}(\mathcal{G}_\text{gene}) \nonumber \\
& = \{\hat{v}_1,\hat{v}_2,\dots,\hat{v}_{N_\text{gene}}\} \in \R^{N_\text{gene} \times d}.   
\end{align}
More complicated gene graph structures that incorporate external knowledge bases, e.g., InBioMap \cite{li2017scored}, can also be applied to the graph-in-graph network as well. However, we mainly aim to demonstrate the effectiveness of capturing the structural information with the gene network for generating GO term description, so we use the simpler encoder formulation.

\subsubsection{Term Graph}
We construct a term graph $\mathcal{G}_\text{term} = (V', E')$ to model biological relationships among terms. $V' = \{v'_i\}_{i=1:N_\text{term}} \in \R^{N_\text{term} \times d}$ is a set of nodes and $E' = \{e'_{i,j}\}_{i,j=1:N_\text{term}}$ is a set of edges. For any two terms, if they have a connection in GO, we add a corresponding edge in $E'$. We represent each node in $V'$ with the embedding $v'_i \in \R^{d}$.
In our implementation, we initialize the embedding $v'_i$ with its corresponding embedding of term node from the gene graph. The edge weights are all set to be 1.

Next, we introduce the term-graph encoder, which is based on the graph convolution operation \cite{Li2016GCN}, to encode the term graph as
\begin{equation}
\hat{v}'_{i}={v}'_{i}+\text{ReLU}\left(\sum\nolimits_{j=1}^{N_\text{term}} {e}'_{i, j} {W}'_{v} {v}'_{j}\right) ,
\end{equation}
where $W'_v \in \R^{d \times d}$ is a learnable matrix.
As a result, we can acquire a set of encoded term node embeddings:
\begin{align}
\label{eqn:term_graph}
\hat{V}' &= \text{Term-Graph-Encoder}(\mathcal{G}_\text{term}) \nonumber \\
& = \{\hat{v}'_1,\hat{v}'_2,\dots,\hat{v}'_{N_\text{term}}\} \in \R^{N_\text{term} \times d}.   
\end{align}
Finally, we concatenate the encoded gene node embeddings $\hat{V}$ from Eq.~(\ref{eqn:gene_graph}) and the encoded term node embeddings $\hat{V}'$ from Eq.~(\ref{eqn:term_graph}) to produce the intermediate representations $I = [\hat{V}; \hat{V}'] \in \R^{(N_\text{gene} + N_\text{term}) \times d}$ for the input term $T$, and then feed these representations into baseline decoders to generate an accurate and coherent GO term description.

\subsection{Description Decoder}
\label{sec:decoder}

As shown in Eq.~(\ref{eq:formulation}), the description decoder aims to generate the final textual description based on the encoded  intermediate representation $I \in \R^{(N_\text{gene} + N_\text{term}) \times d}$.
In implementation, we can choose either the LSTM \cite{Hochreiter1997LSTM} or Transformer \cite{Vaswani2017Transformer} as the decoder.
Specifically, Transformer, which includes the Multi-Head Attention (MHA) and the Feed-Forward Network (FFN), achieves the state-of-the-art performances on multiple natural language generation tasks.
Taking the Transformer decoder as example: for each decoding step $t$, the decoder takes the embedding of the current input word $x_t = w_t + e_t \in \R^{d}$ as input, where $w_t$ and $e_t$ denote the word embedding and fixed position embedding, respectively; we then generate each word $y_t$ in the target description $S = \{y_1, y_2,\dots, y_{N_\text{S}}\}$, which is defined as follows:
\begin{align}
&h_t = \text{MHA}(x_t, x_{1:t}, x_{1:t}) \label{eq:MHA} \\
&h'_t = \text{MHA}(h_t, I, I) \\
&y_{t} \sim p_{t} =\text{softmax}(\text{FFN}(h'_t)\text{W}_p + \text{b}_p) \label{eq:FFN} ,
\end{align}
where $\text{W}_p \in \R^{d \times |D|}$ and $\text{b}_p$ are the learnable parameters ($|D|$: vocabulary size); the MHA and FFN stand for the Multi-Head Attention and Feed-Forward Network in the original Transformer \cite{Vaswani2017Transformer}, respectively. 
In detail, the MHA consists of $n$ parallel heads and each head is defined as a scaled dot-product attention:
\begin{align}
&\text{Att}_i(Q,K,V) = \text{softmax}\left(\frac{Q\text{W}_i^\text{Q}(K\text{W}_i^\text{K})^T}{\sqrt{{d}_{n}}}\right)V\text{W}_i^\text{V} \nonumber \\
& \text{MHA}(Q,K,V) = [\text{Att}_1(Q,K,V); \dots; \text{Att}_n(Q,K,V)]\text{W}^\text{O} ,
\end{align}
where $Q \in \R^{l_Q \times d}$, $K \in \R^{l_K \times d}$, and $V \in \R^{l_V \times d}$ denote the Query matrix, the Key matrix, and the Value matrix, respectively; $\text{W}_i^\text{Q}, \text{W}_i^\text{K}, \text{W}_i^\text{V} \in \R^{d \times d_n}$ and $\text{W}^\text{O} \in \R^{d \times d}$ are learnable parameters, where ${d}_{n} = d / {n}$; $[\cdot;\cdot]$ stands for concatenation operation.

Following the MHA is the FFN, defined as follows:
\begin{align}
\text{FFN}(x) = \max(0,x\text{W}_\text{f}+\text{b}_\text{f})\text{W}_\text{ff}+\text{b}_\text{ff}   
\end{align}
where $\max(0,*)$ represents the ReLU activation function; $\text{W}_\text{f} \in \R^{d \times 4d}$ and $\text{W}_\text{ff} \in \R^{4d \times d}$ denote  learnable matrices for linear transformation; $\text{b}_\text{f}$ and $\text{b}_\text{ff}$ represent the bias terms. It is worth noting that both the MHA and FFN are followed by an operation sequence of dropout \cite{srivastava2014dropout}, residual connection \cite{he2016deep}, and layer normalization \cite{ba2016layernormalization}.
In particular, if LSTM \cite{Hochreiter1997LSTM} is adopted as the description decoder, we can directly replace the MHA in Eq.~(\ref{eq:MHA}) with the LSTM unit and remove the FFN in Eq.~(\ref{eq:FFN}).

Overall, as our Graph-in-Graph network substitutes the original intermediate representations of input term with $I = [\hat{V}; \hat{V}']$, our approach can use the sequence-to-sequence model decoders without any alterations to experimental settings or training strategies.
In our subsequent experiments, we show that our approach can benefit a wide range of downstream sequence-to-sequence models.

\begin{table}[t]
\centering
\caption{The statistics of the three datasets that we constructed, w.r.t. the numbers of terms and genes, and the mean length of descriptions.}
\label{tab:statistics}
\setlength{\tabcolsep}{2.5pt}   
\begin{tabular}{@{}l c c c@{}}
\toprule 

\multirow{2}{*}{Datasets} & \multirow{2}{*}{\begin{tabular}[c]{@{}c@{}} Molecular \\ Function \end{tabular}} & \multirow{2}{*}{\begin{tabular}[c]{@{}c@{}} Biological \\ Process \end{tabular}} & \multirow{2}{*}{\begin{tabular}[c]{@{}c@{}} Cellular \\ Component \end{tabular}} \\ \\
\midrule [\heavyrulewidth]
Number of Terms & 12,257 & 30,490 & 4,463 \\
Number of Genes & 32,002 & 32,840 & 32,477\\
Mean Description Length & 19.29 & 27.49 & 29.36\\
\bottomrule
\end{tabular}
\end{table}

\begin{table*}[t]
\centering
\caption{Performance on our built three benchmark datasets. Higher is better in all columns. We conducted 5 runs with different seeds for all experiments, the t-tests indicate that $p$ $<$ 0.01. As we can see, all the baseline models with significantly different  structures enjoy a comfortable improvement with our Graph-in-Graph network.}
\label{tab:results}
\setlength{\tabcolsep}{5pt}
\begin{tabular}{@{}l c c c c c c c c c@{}}
\toprule 

\multirow{2}{*}[-3pt]{Methods} & \multicolumn{3}{c}{Dataset: Molecular Function} & \multicolumn{3}{c}{Dataset: Biological Process} & \multicolumn{3}{c}{Dataset: Cellular Component}  \\ \cmidrule(lr){2-4} \cmidrule(lr){5-7} \cmidrule(lr){8-10}
& METEOR & ROUGE-L & BLEU & METEOR & ROUGE-L & BLEU & METEOR & ROUGE-L & BLEU  \\
\midrule [\heavyrulewidth]

Seq2Seq \cite{bahdanau2014neural} &  20.5 & 44.7 & 26.3 & 15.1 & 36.8 & 17.6 & 12.9 & 28.3 & 11.7  \\  

\ with Graph-in-Graph & \bf 27.5 & \bf 51.2 & \bf 34.8 & \bf 21.0 & \bf 40.3 & \bf 23.7 & \bf 15.3 & \bf 31.6 & \bf 14.6 \\ \midrule 

GlobalAtt. \cite{Luong2015Seq2Seq_Attention} & 21.9 & 45.5 & 26.8 & 16.0 & 37.3 & 18.1 & 14.5 & 28.9 & 11.8 \\  

\ with Graph-in-Graph & \bf 27.9 & \bf 49.8 & \bf 34.6 & \bf 20.5 & \bf 39.2 & \bf 21.4 & \bf 16.7 & \bf 32.1 &\bf 14.5  \\  \midrule 

HRNN \cite{Lin2015HRNN} & 23.1 & 44.9 & 27.2 & 18.3 & 38.7 & 19.7 & 16.1 & 30.4 & 13.3 \\

\ with Graph-in-Graph & \bf 28.6 & \bf 49.6 & \bf 34.4 & \bf  22.4 & \bf 40.5 & \bf  24.8 & \bf 17.3 & \bf  34.3 & \bf 15.9 \\ \midrule  

CopyNet \cite{Gu2016CopyNet} & 22.0 & 44.5 & 27.0 & 16.4 & 37.7 & 18.5 & 14.7 & 29.2 & 12.3 \\ 

\ with Graph-in-Graph & \bf 25.2 & \bf 47.3 & \bf 31.3 & \bf 19.7 & \bf 39.4 & \bf 21.3 & \bf 15.5 & \bf 30.5 & \bf 13.0 \\ \midrule

ConvS2S \cite{Gehring2017ConvS2S} & 15.7 & 34.1 & 16.2 & 11.4 & 27.5 & 10.5 & 10.2 & 23.9 & 11.1 \\

\ with Graph-in-Graph & \bf 17.5 & \bf 34.8 & \bf 19.7 & \bf 12.5 & \bf 27.9 & \bf 12.8 & \bf 11.6 & \bf 25.0 & \bf 12.7 \\ \midrule  

GraphWriter \cite{Kedziorski2019Text} & 25.3 & 47.2 & 31.7 &  22.3 & 43.5 & 24.8  &  17.0 &  30.9 & 14.1 \\  

\ with Graph-in-Graph  & \bf 27.1 & \bf 52.0 & \bf 36.9 & \bf 25.2 & \bf 45.7 & \bf 26.7 & \bf 17.8 & \bf 32.5 & \bf 17.4 \\   \midrule

Transformer \cite{Vaswani2017Transformer} & 26.7 & 49.6 & 33.5 &  20.9 & 41.6 &  23.4 &  16.1 &  31.4 & 15.2 \\  

\ with Graph-in-Graph & \bf 30.4 & \bf 56.3 & \bf 38.3 & \bf 24.5 & \bf 46.1 & \bf 27.0 & \bf 17.5 & \bf 33.7 & \bf 17.9 \\

\bottomrule
\end{tabular}
\end{table*}

\section{Experiments}

In this section, we first introduce the three datasets that we constructed for experiments, as well as the metrics, base models and settings that we tested. Then, we present the results of our proposed Graph-in-Graph Network.

\subsection{Datasets, Metrics, Baselines and Settings}
\subsubsection{Datasets}
To address the GO description problem, we propose three large-scale benchmark datasets containing the GO terms about \textit{Homo sapiens} (humans), i.e., 1) \textit{Molecular Function}\footnote{\url{http://www.informatics.jax.org/vocab/gene_ontology/GO:0003674}}, 2) \textit{Biological Process}\footnote{\url{http://www.informatics.jax.org/vocab/gene_ontology/GO:0008150}} and 3) \textit{Cellular Component}\footnote{\url{http://www.informatics.jax.org/vocab/gene_ontology/GO:0005575}}, which contain a large number of terms.
Next, following \cite{Zhang2020GO}, we collect the gene text from GeneCards\footnote{\url{https://www.genecards.org/}}\cite{stelzer2016genecards}, which contains the information from Universal Protein Resource (UniProt)\footnote{\url{https://www.uniprot.org/}} \cite{UniProt}. These databases are free for academic research and academic non-profit institutions.
Specifically, the \textit{Creative Commons Attribution (CC BY 4.0) License} (\url{https://www.uniprot.org/help/license}) is used.
The dataset statistics are provided in Table~\ref{tab:statistics}, with the number of terms, the number of genes and the average length of descriptions.
Then, we randomly split the three datasets into 70\%-10\%-20\% train-validation-test splits.
As a result, the Molecular Function/Biological Process/Cellular Component datasets are split into 8,580/21,343/3,124 term-description pairs for training, 1,225/3,049/446 term-description pairs for validation and 2,450/6,098/892 term-description pairs for testing.
Next, we preprocess the descriptions by tokenizing and converting to lower-cases. At last, we filter tokens that occur less than 3 times in the corpus, resulting in a vocabulary of around 5k, 12k, 4k tokens for Molecular Function, Biological Process and Cellular Component datasets, respectively.

\subsubsection{Metrics}
In our experiment, we evaluate the performance of models on the widely-used natural language generation metrics, i.e., BLEU \cite{papineni2002bleu}, METEOR \cite{Banerjee2005METEOR} and ROUGE-L \cite{lin2004rouge}, which are reported by the evaluation toolkit \cite{chen2015microsoft}\footnote{\url{https://github.com/tylin/coco-caption}}.
In particular, BLEU and METEOR were originally designed for machine translation evaluation. ROUGE-L automatically evaluates extracted text summarization.

\subsubsection{Baselines} 
In our experiments, we choose seven sequence-to-sequence models with different structures as baseline models, i.e., 1) RNN-based model (\textbf{Seq2Seq}) \cite{bahdanau2014neural}, 2) attention-based model (\textbf{GlobalAtt.}) \cite{Luong2015Seq2Seq_Attention}, 3) hierarchical RNN-based model (\textbf{HRNN}) \cite{Lin2015HRNN}, 4) copy mechanism based model (\textbf{CopyNet}) \cite{Gu2016CopyNet}, 5) convolutional-neural-network- (CNN-) based model (\textbf{ConvS2S}) \cite{Gehring2017ConvS2S}, 6) graph-to-sequence model (\textbf{GraphWriter}) \cite{Kedziorski2019Text}, and 7) fully-attentive model (\textbf{Transformer}) \cite{Vaswani2017Transformer}, which are widely used in current sequence-to-sequence tasks, e.g., neural machine translation and paraphrasing. In detail, to demonstrate the effectiveness of the proposed graph-in-graph network, we compare the performance of the models with and without the graph-in-graph network.

\subsubsection{Settings}

We use the embedding size $d=512$ for both the graph-in-graph network and the baseline models. 
To obtain the parent and child nodes of the current term of our graph-in-graph network, we retrieve the term nodes, whose genes cover all genes in current term, as the parent nodes, and the term nodes, whose genes are a subset of current term, as the child nodes. 
We use the Adam optimizer~\cite{kingma2014adam} with a batch size of 16 and a learning rate of 1e-3 for parameter optimization.

Since our proposed graph-in-graph is regarded as a pluggable module to explore the relations between terms and the relations between genes, we keep the inner structure of each of the baselines unchanged and maintain the same parameter initialization and training strategy.
Specifically, as shown in the GO Description Problem Definition (section~\ref{sec:definition}) and Equation~\ref{eq:definition} of our Approach section, we take the input term $T$, consisting of $N_\text{G}$ genes (vectors) as a sequence, and adopt the sequence-to-sequence baselines to generate a description to describe the biological function of the input term.
To incorporate our graph-in-graph into baselines, as shown in Equation~\ref{eq:formulation}, our approach can produce the $\mathcal{G}_\text{term}$ and the $\mathcal{G}_\text{gene}$ by capturing the biological relations between terms and the gene relations within each term, respectively. 
Next, as shown in the last paragraph of Section~\ref{sec:decoder}, we substitute the original intermediate representations $I$ of input term $T$ with $[\mathcal{G}_\text{term}; \mathcal{G}_\text{gene}]$, i.e., $I = [\hat{V}; \hat{V}']$, so our approach can use the sequence-to-sequence model decoders without any alterations to experimental settings or training strategies.
All re-implementations and our experiments were run on 4 V100 GPUs for approximately seven days.

\subsection{Results}
The experimental results on our built three benchmark datasets are reported in Table~\ref{tab:results}. As shown, our proposed graph-in-graph network can consistently boost all baselines across all metrics, with a relative BLEU score improvement of 14.3\%$\sim$32.3\%, 15.1\%$\sim$34.7\%, and 5.7\%$\sim$24.8\% for Molecular Function, Biological Process, and Cellular Component, respectively, where the Seq2Seq with Graph-in-Graph achieves the greatest improvements. 
The results prove the effectiveness of our approach in exploring the relationships between terms and the relationships between genes.

\begin{table*}[t]
\centering
\caption{Ablation study of our approach, which includes the Gene Graph and Term Graph encoders, on the Molecular Function, Biological Process, and Cellular Component datasets. PN and CN denote the Parent Nodes and Child Nodes, respectively. Higher scores are better in all columns. Full Model represents the baseline model with the Graph-in-Graph. As we can see, each component of our approach boosts the performances of the baseline models across all metrics.}
\label{tab:ablation}
\setlength{\tabcolsep}{4pt}   
\begin{tabular}{@{}c c c c c c c c c c c c c@{}}
\toprule 

\multirow{2}{*}[-3pt]{Methods} & \multirow{2}{*}[-2pt]{\begin{tabular}[c]{@{}c@{}} Gene  \\ Graph \end{tabular}} &  \multicolumn{2}{c}{Term Graph} & \multicolumn{3}{c}{Dataset: Molecular Function} & \multicolumn{3}{c}{Dataset: Biological Process} & \multicolumn{3}{c}{Dataset: Cellular Component}  \\ \cmidrule(lr){3-4} \cmidrule(lr){5-7} \cmidrule(lr){8-10} \cmidrule(lr){11-13}
& & PN & CN & METEOR & ROUGE-L & BLEU & METEOR & ROUGE-L & BLEU & METEOR & ROUGE-L & BLEU \\
\midrule [\heavyrulewidth]

Seq2Seq  & & & &  20.5 & 44.7 & 26.3 & 15.1 & 36.8 & 17.6 & 12.9 & 28.3 & 11.7  \\    

(a) & $\surd$  & & & 23.6 & 46.3 & 29.7 & 16.5 & 37.1 & 18.4 & 13.7 & 29.0 & 12.5 \\

(b) & $\surd$ & $\surd$ & & 24.1 & 47.5 & 31.0 & 17.5 & 37.9 & 19.5 & 14.2 & 29.1 & 12.7 \\

(c) & $\surd$ & & $\surd$ & 27.2 & 50.8 & 34.0 & 20.3 & 39.4 & 22.9 & 15.1 & 30.8 & 14.2 \\

Full Model & $\surd$ & $\surd$ & $\surd$ & \bf 27.5 & \bf 51.2 & \bf 34.8 & \bf 21.0 & \bf 40.3 & \bf 23.7 & \bf 15.3 & \bf 31.6 & \bf 14.6 \\

\midrule 

Transformer &  &  &  & 26.7 & 49.6 & 33.5 &  20.9 & 41.6 &  23.4 &  16.1 &  31.4 & 15.2 \\  

(a) & $\surd$ &  & & 27.3 & 51.2 & 34.4 & 21.8 & 42.9 & 25.0 & 16.6 & 32.1 & 16.3 \\

(b) & $\surd$ & $\surd$ & & 27.5 & 52.3 & 35.1 & 22.0 & 43.5 & 25.3 & 16.4 & 32.1 & 15.8 \\

(c) & $\surd$ & & $\surd$ & 29.8 & 55.7 & 38.0 & 24.1 & 45.3 & 26.2 & 17.3 & 33.3 & 17.5 \\

Full Model & $\surd$ & $\surd$ & $\surd$ & \bf 30.4 & \bf 56.3 & \bf 38.3 & \bf 24.5 & \bf 46.1 & \bf 27.0 & \bf 17.5 & \bf 33.7 & \bf 17.9 \\

\bottomrule
\end{tabular}
\end{table*}

\subsection{Ablation Study}
We select two mainstream sequence-to-sequence models, i.e., Seq2Seq and Transformer, where the latter achieves state-of-the-art performances on multiple tasks, to evaluate the contribution of each proposed module, i.e., Gene Graph and Term Graph encoders (Table~\ref{tab:ablation}). As we can see, each component in our proposed approach can boost the performances of baselines over all metrics, verifying the effectiveness of our approach. 

\subsubsection{Effect of Gene Graph Encoder} \ \ In particular, setting (a) in Table~\ref{tab:ablation} shows that the Gene Graph encoder can successfully boost baselines with relative gains up to 12.9\%, 10.7\% and 7.2\% for Molecular Function, Biological Process and Cellular Component datasets in terms of BLEU score respectively, demonstrating how our proposed Gene Graph exploits the relations between genes to promote the performance.

\subsubsection{Effect of Term Graph Encoder} \ \ Settings (b,c) show that modeling either parent nodes or child nodes respectively both boost performance, indicating the importance of introducing enriched information from parent and child nodes into the models. 
In particular, since the genes of the current node are a subset of the parent nodes and the genes of the child nodes are a subset of the current node, modeling parent nodes encourages the model to utilize the abstractive information from parents, and modeling children encourages the model to summarize the information from child nodes, similar to the task of text summarization.
Moreover, by comparing the results of (b) and (c), we observe that the child nodes introduce more improvements than the parent nodes. We attribute this to the fact that child nodes contain more specific textual information than the parent nodes, which we verify in our following quantitative analysis. Since the Gene Graph encoder and Term Graph encoder can improve the performance from different information sources, combining them can lead to the most prominent improvement across all metrics (see Full Model), with up to 32.3\%, 34.7\%, 24.8\% BLEU score improvement for the Molecular Function, Biological Process, Cellular Component datasets, respectively.

\begin{table*}[t]
\centering
\caption{Out-of-Domain analysis on the Transformer (Baseline) and the Transformer with Graph-in-Graph model (Ours). The principal diagonal reports the in-domain results, and the off-diagonal reports the out-of-domain results. The ($\downarrow$ Number) denotes the decreased performance of out-of-domain results compared to in-domain results.}
\label{tab:ood}
\setlength{\tabcolsep}{3pt}   
\begin{tabular}{@{}l l c c c|c c c|c c c@{}}
\toprule

\multirow{2}{*}[-2pt]{\begin{tabular}[c]{@{}l@{}} Training  \\ Dataset \end{tabular}} & \multirow{2}{*}[-3pt]{Methods} & \multicolumn{3}{c|}{Dataset: Molecular Function} & \multicolumn{3}{c|}{Dataset: Biological Process} & \multicolumn{3}{c}{Dataset: Cellular Component}  \\ \cmidrule(lr){3-5} \cmidrule(lr){6-8} \cmidrule(lr){9-11}
& & METEOR & ROUGE-L & BLEU & METEOR & ROUGE-L & BLEU & METEOR & ROUGE-L & BLEU \\
\midrule [\heavyrulewidth]

\multirow{2}{*}{\begin{tabular}[c]{@{}l@{}} Molecular  \\ Function \end{tabular}} & Baseline & 26.7 ($-$) & 49.6 ($-$) & 33.5 ($-$) & 3.5 ($\downarrow$17.4) & 4.1 ($\downarrow$37.5) & 1.3 ($\downarrow$22.1) & 5.8 ($\downarrow$10.3) & 11.3 ($\downarrow$20.1) & 2.7 ($\downarrow$12.5)  \\  

& Ours & \textbf{30.4} ($-$) &\textbf{56.3} ($-$) & \textbf{38.3} ($-$) & \textbf{4.4} ($\downarrow$20.1) & \textbf{5.8} ($\downarrow$40.3) & \textbf{2.9} ($\downarrow$24.1) & \textbf{6.7} ($\downarrow$10.8) & \textbf{15.0} ($\downarrow$18.7) & \textbf{5.9} ($\downarrow$12.0)  \\ \midrule

\multirow{2}{*}{\begin{tabular}[c]{@{}l@{}} Biological  \\ Process \end{tabular}} & Baseline & 14.0 ($\downarrow$12.7) & 25.5 ($\downarrow$24.1) & 13.5 ($\downarrow$20.0) & 20.9 ($-$) & 41.6 ($-$) & 23.4 ($-$) & 11.8 ($\downarrow$4.3) & 22.2 ($\downarrow$9.2) & 10.3 ($\downarrow$4.9)\\

& Ours & \textbf{20.3} ($\downarrow$10.1) & \textbf{37.1} ($\downarrow$19.2) & \textbf{25.0} ($\downarrow$13.3) & \textbf{24.5} ($-$) & \textbf{46.1} ($-$) & \textbf{27.0} ($-$) & \textbf{15.6} ($\downarrow$1.9) & \textbf{28.8} ($\downarrow$4.9) & \textbf{15.5} ($\downarrow$2.4) \\ \midrule

\multirow{2}{*}{\begin{tabular}[c]{@{}l@{}} Cellular  \\ Component \end{tabular}} & Baseline  & \ \ 7.7 ($\downarrow$19.0) & 12.4 ($\downarrow$37.2) & \ \ 4.8 ($\downarrow$28.7) & 10.7 ($\downarrow$10.2) & 15.3 ($\downarrow$26.3) & 7.5 ($\downarrow$15.9) & 16.1  ($-$) & 31.4  ($-$)& 15.2 ($-$)  \\

& Ours & \textbf{12.6} ($\downarrow$17.8) & \textbf{17.6} ($\downarrow$38.7) & \textbf{10.2} ($\downarrow$28.1) & \textbf{13.4} ($\downarrow$11.1) & \textbf{21.5} ($\downarrow$24.6) & \textbf{9.9} ($\downarrow$17.1) & \textbf{17.5} ($-$) & \textbf{33.7} ($-$) & \textbf{17.9} ($-$) \\ 

\bottomrule
\end{tabular}
\end{table*}

\subsection{Out-of-Domain Analysis}
In this section, we further conduct an out-of-domain analysis to examine whether our model can aid in the discovery of new Gene Ontology terms through describing the function of terms from \textbf{a new domain}. 
Specifically, we perform a cross-dataset prediction by training on one dataset and evaluating on the other two datasets. We summarized the performance of our method and base models in Table~\ref{tab:ood}. As we show, our approach consistently outperforms the baseline methods. Although we observe that the performance of the baseline models and our approach all decrease on the out-of-domain datasets, our approach achieves a lower drop in performance compared to the baseline models on most cases, which further demonstrates the effectiveness and the transferability of our approach. 
More encouragingly, when transferring our approach trained on Biological Process dataset into Cellular Component dataset, we observed a superior out-of-domain performance when using our method, i.e., METEOR/ROUGE-L/BLEU: 15.6/28.8/15.5, which are competitive with the in-domain results of baseline, i.e., METEOR/ROUGE-L/BLEU: 16.1/31.4/15.2.
The superior performance on the out-of-domain datasets demonstrate the effectiveness of our approach with boosted transferability of models, leading to higher quality descriptions for new GO terms from a new domain than baseline models.

\subsection{Qualitative Analysis}
\begin{figure*}[t]

\centering
\includegraphics[width=1\linewidth]{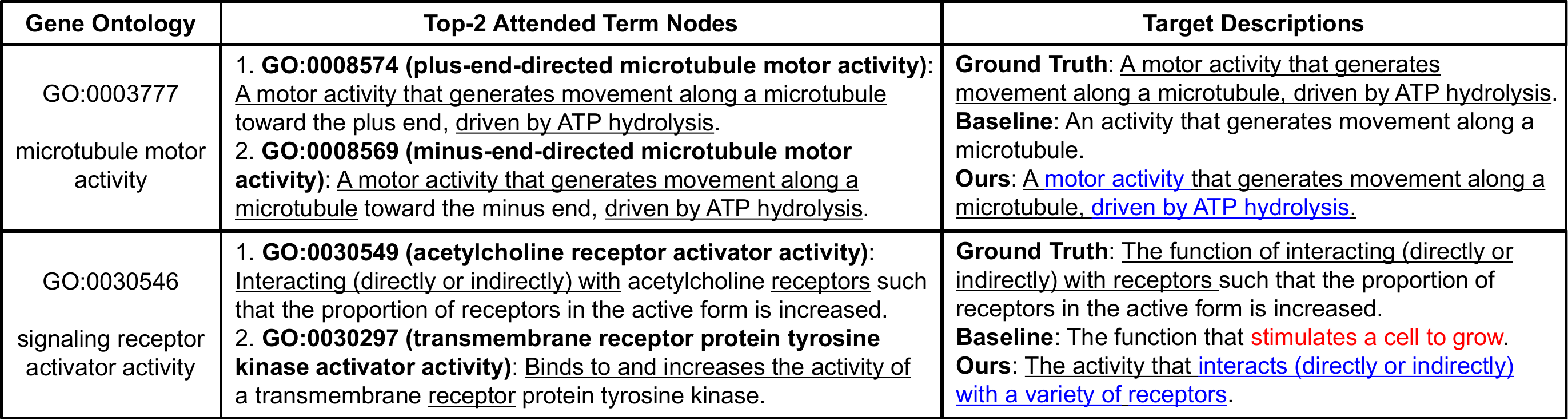}
\caption{Examples of the generated descriptions of GO from baselines and from our approach (i.e., baselines with Graph-in-Graph). We also visualize the attended term nodes with top-2 attention weights in the attention mechanism of the decoder and their corresponding descriptions. Please view in color. The Blue colored text denotes the examples when our approach generates better descriptions than the baseline, while Red denotes unfavorable descriptions. Underlined text denotes alignment between the ground truth text and generated/attended text.}
\label{fig:example}
\end{figure*}

In Figure~\ref{fig:example}, we conduct a qualitative analysis to more thoroughly understand our approach.
We can see that the descriptions generated by our proposed method are better aligned with ground truth descriptions than baseline models.
For example, our approach correctly generates the key gene function ``\textit{motor activity}'' and ``\textit{driven by ATP hydrolysis}'' in the first example and ``\textit{interacts (directly or indirectly) with a variety of receptors}'' in the second example.
We attribute this to the capability of our Graph-in-Graph framework to explore the structural information of terms and then exploit the abstract information from parent nodes and the detailed information from child nodes to improve the description generation. This can be verified by the strong overlap between the generated description and the description of attended nodes.
The qualitative analysis further supports our hypothesis and verifies the effectiveness of our proposed approach in exploring the structural information of GO to boost the performance of description generation.

\section{Conclusion and Discussion}

In this paper, we make the first attempt to automatically describe the function of a Gene Ontology term. We propose the novel Graph-in-Graph framework, which introduces a Gene Graph encoder and a Term Graph encoder to explore the structural information of the relationships between genes and the relationships between terms, respectively. The extensive experiments and analyses on our three benchmark datasets verify the effectiveness and the generalization capabilities of our approach, which consistently demonstrates superior performance on a wide range of baseline models with substantially different model structures, across all metrics. The greatest relative BLEU, ROUGE-L, and METEOR score improvements of our method are 34.7\%, 14.5\%, and 39.1\%, respectively. The analysis further proves the transferability of our approach.

Although this paper makes the first attempt to automatically describe the biological function of Gene Ontology (GO) terms, which can assist biologists in GO construction and analysis, the training of our proposed model relies on a large volume of GO-description pairs. Such limitations may have following consequences: 1) The performance will be limited by the size of existing GO terms in biology domain, but could be alleviated using techniques such as distillations from publicly-available pre-trained models, e.g., BioBERT \cite{BioBERT}; 2) When applying to new biology domains, we need to collect a new set of GO descriptions, which may be difficult and time-consuming for some biology domains \cite{wang2022textomics,wang2021leveraging}.

In the future, it can be interesting to apply the Graph-in-Graph network to improve other medical tasks.
Take drug mining and recommendation for example, the inner graph may refer to the molecular structure of drugs and the outer graph may refer to the knowledge graph that connects drugs, diseases, symptoms and other medical entities.

\section*{Acknowledgments}
We thank all the anonymous reviewers for their constructive comments and suggestions.

\bibliographystyle{ACM-Reference-Format}
\bibliography{sample}


\begin{thebibliography}{46}


\ifx \showCODEN    \undefined \def \showCODEN     #1{\unskip}     \fi
\ifx \showDOI      \undefined \def \showDOI       #1{#1}\fi
\ifx \showISBNx    \undefined \def \showISBNx     #1{\unskip}     \fi
\ifx \showISBNxiii \undefined \def \showISBNxiii  #1{\unskip}     \fi
\ifx \showISSN     \undefined \def \showISSN      #1{\unskip}     \fi
\ifx \showLCCN     \undefined \def \showLCCN      #1{\unskip}     \fi
\ifx \shownote     \undefined \def \shownote      #1{#1}          \fi
\ifx \showarticletitle \undefined \def \showarticletitle #1{#1}   \fi
\ifx \showURL      \undefined \def \showURL       {\relax}        \fi
\providecommand\bibfield[2]{#2}
\providecommand\bibinfo[2]{#2}
\providecommand\natexlab[1]{#1}
\providecommand\showeprint[2][]{arXiv:#2}

\bibitem[Ashburner et~al\mbox{.}(2000)]%
        {ashburner2000gene}
\bibfield{author}{\bibinfo{person}{Michael Ashburner},
  \bibinfo{person}{Catherine~A Ball}, \bibinfo{person}{Judith~A Blake},
  \bibinfo{person}{David Botstein}, \bibinfo{person}{Heather Butler},
  \bibinfo{person}{J~Michael Cherry}, \bibinfo{person}{Allan~P Davis},
  \bibinfo{person}{Kara Dolinski}, \bibinfo{person}{Selina~S Dwight},
  \bibinfo{person}{Janan~T Eppig}, {et~al\mbox{.}}}
  \bibinfo{year}{2000}\natexlab{}.
\newblock \showarticletitle{Gene Ontology: Tool for the Unification of
  Biology}.
\newblock \bibinfo{journal}{\emph{Nature genetics}} \bibinfo{volume}{25},
  \bibinfo{number}{1} (\bibinfo{year}{2000}), \bibinfo{pages}{25--29}.
\newblock
\urldef\tempurl%
\url{https://doi.org/10.1038/75556}
\showURL{%
\tempurl}


\bibitem[Ba et~al\mbox{.}(2016)]%
        {ba2016layernormalization}
\bibfield{author}{\bibinfo{person}{Lei~Jimmy Ba}, \bibinfo{person}{Jamie~Ryan
  Kiros}, {and} \bibinfo{person}{Geoffrey~E. Hinton}.}
  \bibinfo{year}{2016}\natexlab{}.
\newblock \showarticletitle{Layer Normalization}.
\newblock \bibinfo{journal}{\emph{CoRR}}  \bibinfo{volume}{abs/1607.06450}
  (\bibinfo{year}{2016}).
\newblock
\showeprint[arXiv]{1607.06450}
\urldef\tempurl%
\url{http://arxiv.org/abs/1607.06450}
\showURL{%
\tempurl}


\bibitem[Bahdanau et~al\mbox{.}(2015)]%
        {bahdanau2014neural}
\bibfield{author}{\bibinfo{person}{Dzmitry Bahdanau},
  \bibinfo{person}{Kyunghyun Cho}, {and} \bibinfo{person}{Yoshua Bengio}.}
  \bibinfo{year}{2015}\natexlab{}.
\newblock \showarticletitle{Neural Machine Translation by Jointly Learning to
  Align and Translate}. In \bibinfo{booktitle}{\emph{3rd International
  Conference on Learning Representations, {ICLR} 2015}}.
  \bibinfo{publisher}{ICLR}, \bibinfo{address}{San Diego, CA, USA}.
\newblock
\urldef\tempurl%
\url{http://arxiv.org/abs/1409.0473}
\showURL{%
\tempurl}


\bibitem[Banerjee and Lavie(2005)]%
        {Banerjee2005METEOR}
\bibfield{author}{\bibinfo{person}{Satanjeev Banerjee} {and}
  \bibinfo{person}{Alon Lavie}.} \bibinfo{year}{2005}\natexlab{}.
\newblock \showarticletitle{{METEOR:} An Automatic Metric for {MT} Evaluation
  with Improved Correlation with Human Judgments}. In
  \bibinfo{booktitle}{\emph{Proceedings of the Workshop on Intrinsic and
  Extrinsic Evaluation Measures for Machine Translation and/or
  Summarization@ACL 2005}}. \bibinfo{publisher}{Association for Computational
  Linguistics}, \bibinfo{address}{Ann Arbor, Michigan, USA},
  \bibinfo{pages}{65--72}.
\newblock
\urldef\tempurl%
\url{https://aclanthology.org/W05-0909/}
\showURL{%
\tempurl}


\bibitem[Cai and Lam(2020)]%
        {Cai2020Graph}
\bibfield{author}{\bibinfo{person}{Deng Cai} {and} \bibinfo{person}{Wai Lam}.}
  \bibinfo{year}{2020}\natexlab{}.
\newblock \showarticletitle{Graph Transformer for Graph-to-Sequence Learning}.
  In \bibinfo{booktitle}{\emph{The Thirty-Fourth {AAAI} Conference on
  Artificial Intelligence, {AAAI} 2020, The Thirty-Second Innovative
  Applications of Artificial Intelligence Conference, {IAAI} 2020, The Tenth
  {AAAI} Symposium on Educational Advances in Artificial Intelligence, {EAAI}
  2020}}. \bibinfo{publisher}{{AAAI} Press}, \bibinfo{address}{New York, NY,
  USA}, \bibinfo{pages}{7464--7471}.
\newblock
\urldef\tempurl%
\url{https://aaai.org/ojs/index.php/AAAI/article/view/6243}
\showURL{%
\tempurl}


\bibitem[Chen et~al\mbox{.}(2015)]%
        {chen2015microsoft}
\bibfield{author}{\bibinfo{person}{Xinlei Chen}, \bibinfo{person}{Hao Fang},
  \bibinfo{person}{Tsung{-}Yi Lin}, \bibinfo{person}{Ramakrishna Vedantam},
  \bibinfo{person}{Saurabh Gupta}, \bibinfo{person}{Piotr Doll{\'{a}}r}, {and}
  \bibinfo{person}{C.~Lawrence Zitnick}.} \bibinfo{year}{2015}\natexlab{}.
\newblock \showarticletitle{Microsoft {COCO} Captions: Data Collection and
  Evaluation Server}.
\newblock \bibinfo{journal}{\emph{CoRR}}  \bibinfo{volume}{abs/1504.00325}
  (\bibinfo{year}{2015}).
\newblock
\showeprint[arXiv]{1504.00325}
\urldef\tempurl%
\url{http://arxiv.org/abs/1504.00325}
\showURL{%
\tempurl}


\bibitem[Davis et~al\mbox{.}(2016)]%
        {davis2016generating}
\bibfield{author}{\bibinfo{person}{Allan~Peter Davis},
  \bibinfo{person}{Thomas~C Wiegers}, \bibinfo{person}{Benjamin~L King},
  \bibinfo{person}{Jolene Wiegers}, \bibinfo{person}{Cynthia~J Grondin},
  \bibinfo{person}{Daniela Sciaky}, \bibinfo{person}{Robin~J Johnson}, {and}
  \bibinfo{person}{Carolyn~J Mattingly}.} \bibinfo{year}{2016}\natexlab{}.
\newblock \showarticletitle{Generating Gene Ontology-disease Inferences to
  Explore Mechanisms of Human Disease at the Comparative Toxicogenomics
  Database}.
\newblock \bibinfo{journal}{\emph{PLoS One}} \bibinfo{volume}{11},
  \bibinfo{number}{5} (\bibinfo{year}{2016}), \bibinfo{pages}{e0155530}.
\newblock
\urldef\tempurl%
\url{https://journals.plos.org/plosone/article?id=10.1371/journal.pone.0155530}
\showURL{%
\tempurl}


\bibitem[Devlin et~al\mbox{.}(2019)]%
        {Devlin2019BERT}
\bibfield{author}{\bibinfo{person}{Jacob Devlin}, \bibinfo{person}{Ming{-}Wei
  Chang}, \bibinfo{person}{Kenton Lee}, {and} \bibinfo{person}{Kristina
  Toutanova}.} \bibinfo{year}{2019}\natexlab{}.
\newblock \showarticletitle{{BERT:} Pre-training of Deep Bidirectional
  Transformers for Language Understanding}. In
  \bibinfo{booktitle}{\emph{Proceedings of the 2019 Conference of the North
  American Chapter of the Association for Computational Linguistics: Human
  Language Technologies, {NAACL-HLT} 2019}}. \bibinfo{pages}{4171--4186}.
\newblock


\bibitem[Dutkowski et~al\mbox{.}(2013)]%
        {dutkowski2013gene}
\bibfield{author}{\bibinfo{person}{Janusz Dutkowski}, \bibinfo{person}{Michael
  Kramer}, \bibinfo{person}{Michal~A Surma}, \bibinfo{person}{Rama
  Balakrishnan}, \bibinfo{person}{J~Michael Cherry}, \bibinfo{person}{Nevan~J
  Krogan}, {and} \bibinfo{person}{Trey Ideker}.}
  \bibinfo{year}{2013}\natexlab{}.
\newblock \showarticletitle{A Gene Ontology Inferred from Molecular Networks}.
\newblock \bibinfo{journal}{\emph{Nature biotechnology}} \bibinfo{volume}{31},
  \bibinfo{number}{1} (\bibinfo{year}{2013}), \bibinfo{pages}{38--45}.
\newblock
\urldef\tempurl%
\url{https://doi.org/10.1038/nbt.2463}
\showURL{%
\tempurl}


\bibitem[Gehring et~al\mbox{.}(2017)]%
        {Gehring2017ConvS2S}
\bibfield{author}{\bibinfo{person}{Jonas Gehring}, \bibinfo{person}{Michael
  Auli}, \bibinfo{person}{David Grangier}, \bibinfo{person}{Denis Yarats},
  {and} \bibinfo{person}{Yann~N. Dauphin}.} \bibinfo{year}{2017}\natexlab{}.
\newblock \showarticletitle{Convolutional Sequence to Sequence Learning}. In
  \bibinfo{booktitle}{\emph{Proceedings of the 34th International Conference on
  Machine Learning, {ICML} 2017}} \emph{(\bibinfo{series}{Proceedings of
  Machine Learning Research}, Vol.~\bibinfo{volume}{70})}.
  \bibinfo{publisher}{{PMLR}}, \bibinfo{address}{Sydney, NSW, Australia},
  \bibinfo{pages}{1243--1252}.
\newblock
\urldef\tempurl%
\url{http://proceedings.mlr.press/v70/gehring17a.html}
\showURL{%
\tempurl}


\bibitem[Gligorijevic et~al\mbox{.}(2014)]%
        {Gligorijevic2014Integration}
\bibfield{author}{\bibinfo{person}{Vladimir Gligorijevic}, \bibinfo{person}{Vuk
  Janjic}, {and} \bibinfo{person}{Natasa Przulj}.}
  \bibinfo{year}{2014}\natexlab{}.
\newblock \showarticletitle{Integration of molecular network data reconstructs
  Gene Ontology}.
\newblock \bibinfo{journal}{\emph{Bioinformatics}} \bibinfo{volume}{30},
  \bibinfo{number}{17} (\bibinfo{year}{2014}), \bibinfo{pages}{594--600}.
\newblock
\urldef\tempurl%
\url{https://doi.org/10.1093/bioinformatics/btu470}
\showDOI{\tempurl}


\bibitem[Gu et~al\mbox{.}(2016)]%
        {Gu2016CopyNet}
\bibfield{author}{\bibinfo{person}{Jiatao Gu}, \bibinfo{person}{Zhengdong Lu},
  \bibinfo{person}{Hang Li}, {and} \bibinfo{person}{Victor O.~K. Li}.}
  \bibinfo{year}{2016}\natexlab{}.
\newblock \showarticletitle{Incorporating Copying Mechanism in
  Sequence-to-Sequence Learning}. In \bibinfo{booktitle}{\emph{Proceedings of
  the 54th Annual Meeting of the Association for Computational Linguistics
  (Volume 1: Long Papers)}}. \bibinfo{publisher}{Association for Computational
  Linguistics}, \bibinfo{address}{Berlin, Germany},
  \bibinfo{pages}{1631--1640}.
\newblock
\urldef\tempurl%
\url{https://doi.org/10.18653/v1/P16-1154}
\showDOI{\tempurl}


\bibitem[He et~al\mbox{.}(2016)]%
        {he2016deep}
\bibfield{author}{\bibinfo{person}{Kaiming He}, \bibinfo{person}{Xiangyu
  Zhang}, \bibinfo{person}{Shaoqing Ren}, {and} \bibinfo{person}{Jian Sun}.}
  \bibinfo{year}{2016}\natexlab{}.
\newblock \showarticletitle{Deep Residual Learning for Image Recognition}. In
  \bibinfo{booktitle}{\emph{2016 {IEEE} Conference on Computer Vision and
  Pattern Recognition, {CVPR} 2016}}. \bibinfo{publisher}{{IEEE} Computer
  Society}, \bibinfo{address}{Las Vegas, NV, USA}, \bibinfo{pages}{770--778}.
\newblock
\urldef\tempurl%
\url{https://doi.org/10.1109/CVPR.2016.90}
\showDOI{\tempurl}


\bibitem[Hochreiter and Schmidhuber(1997)]%
        {Hochreiter1997LSTM}
\bibfield{author}{\bibinfo{person}{Sepp Hochreiter} {and}
  \bibinfo{person}{J{\"{u}}rgen Schmidhuber}.} \bibinfo{year}{1997}\natexlab{}.
\newblock \showarticletitle{Long Short-Term Memory}.
\newblock \bibinfo{journal}{\emph{Neural Computation}} \bibinfo{volume}{9},
  \bibinfo{number}{8} (\bibinfo{year}{1997}), \bibinfo{pages}{1735--1780}.
\newblock
\urldef\tempurl%
\url{https://doi.org/10.1162/neco.1997.9.8.1735}
\showDOI{\tempurl}


\bibitem[Kingma and Ba(2015)]%
        {kingma2014adam}
\bibfield{author}{\bibinfo{person}{Diederik~P. Kingma} {and}
  \bibinfo{person}{Jimmy Ba}.} \bibinfo{year}{2015}\natexlab{}.
\newblock \showarticletitle{Adam: {A} Method for Stochastic Optimization}. In
  \bibinfo{booktitle}{\emph{3rd International Conference on Learning
  Representations, {ICLR} 2015}}. \bibinfo{publisher}{ICLR},
  \bibinfo{address}{San Diego, CA, USA}.
\newblock
\urldef\tempurl%
\url{http://arxiv.org/abs/1412.6980}
\showURL{%
\tempurl}


\bibitem[Koncel{-}Kedziorski et~al\mbox{.}(2019)]%
        {Kedziorski2019Text}
\bibfield{author}{\bibinfo{person}{Rik Koncel{-}Kedziorski},
  \bibinfo{person}{Dhanush Bekal}, \bibinfo{person}{Yi Luan},
  \bibinfo{person}{Mirella Lapata}, {and} \bibinfo{person}{Hannaneh
  Hajishirzi}.} \bibinfo{year}{2019}\natexlab{}.
\newblock \showarticletitle{Text Generation from Knowledge Graphs with Graph
  Transformers}. In \bibinfo{booktitle}{\emph{Proceedings of the 2019
  Conference of the North American Chapter of the Association for Computational
  Linguistics: Human Language Technologies, {NAACL-HLT} 2019}}.
  \bibinfo{publisher}{Association for Computational Linguistics},
  \bibinfo{address}{Minneapolis, MN, USA}, \bibinfo{pages}{2284--2293}.
\newblock
\urldef\tempurl%
\url{https://doi.org/10.18653/v1/n19-1238}
\showDOI{\tempurl}


\bibitem[Koopmans et~al\mbox{.}(2019)]%
        {koopmans2019syngo}
\bibfield{author}{\bibinfo{person}{Frank Koopmans}, \bibinfo{person}{Pim van
  Nierop}, \bibinfo{person}{Maria Andres-Alonso}, \bibinfo{person}{Andrea
  Byrnes}, \bibinfo{person}{Tony Cijsouw}, \bibinfo{person}{Marcelo~P Coba},
  \bibinfo{person}{L~Niels Cornelisse}, \bibinfo{person}{Ryan~J Farrell},
  \bibinfo{person}{Hana~L Goldschmidt}, \bibinfo{person}{Daniel~P Howrigan},
  {et~al\mbox{.}}} \bibinfo{year}{2019}\natexlab{}.
\newblock \showarticletitle{SynGO: An Evidence-Based, Expert-Curated Knowledge
  Base for the Synapse}.
\newblock \bibinfo{journal}{\emph{Neuron}} \bibinfo{volume}{103},
  \bibinfo{number}{2} (\bibinfo{year}{2019}), \bibinfo{pages}{217--234}.
\newblock
\urldef\tempurl%
\url{https://doi.org/10.1016/j.neuron.2019.05.002}
\showURL{%
\tempurl}


\bibitem[Kramer et~al\mbox{.}(2014)]%
        {Kramer2014Inferring}
\bibfield{author}{\bibinfo{person}{Michael Kramer}, \bibinfo{person}{Janusz
  Dutkowski}, \bibinfo{person}{Michael Yu}, \bibinfo{person}{Vineet Bafna},
  {and} \bibinfo{person}{Trey Ideker}.} \bibinfo{year}{2014}\natexlab{}.
\newblock \showarticletitle{Inferring gene ontologies from pairwise similarity
  data}.
\newblock \bibinfo{journal}{\emph{Bioinformatics}} \bibinfo{volume}{30},
  \bibinfo{number}{12} (\bibinfo{year}{2014}), \bibinfo{pages}{34--42}.
\newblock
\urldef\tempurl%
\url{https://doi.org/10.1093/bioinformatics/btu282}
\showDOI{\tempurl}


\bibitem[Lee et~al\mbox{.}(2020)]%
        {BioBERT}
\bibfield{author}{\bibinfo{person}{Jinhyuk Lee}, \bibinfo{person}{Wonjin Yoon},
  \bibinfo{person}{Sungdong Kim}, \bibinfo{person}{Donghyeon Kim},
  \bibinfo{person}{Sunkyu Kim}, \bibinfo{person}{Chan~Ho So}, {and}
  \bibinfo{person}{Jaewoo Kang}.} \bibinfo{year}{2020}\natexlab{}.
\newblock \showarticletitle{BioBERT: a pre-trained biomedical language
  representation model for biomedical text mining}.
\newblock \bibinfo{journal}{\emph{Bioinformatics}} \bibinfo{volume}{36},
  \bibinfo{number}{4} (\bibinfo{year}{2020}), \bibinfo{pages}{1234--1240}.
\newblock


\bibitem[Li and Yip(2016)]%
        {li2016integrating}
\bibfield{author}{\bibinfo{person}{Le Li} {and} \bibinfo{person}{Kevin~Y Yip}.}
  \bibinfo{year}{2016}\natexlab{}.
\newblock \showarticletitle{Integrating Information in Biological Ontologies
  and Molecular Networks to Infer Novel Terms}.
\newblock \bibinfo{journal}{\emph{Scientific reports}} \bibinfo{volume}{6},
  \bibinfo{number}{1} (\bibinfo{year}{2016}), \bibinfo{pages}{1--10}.
\newblock
\urldef\tempurl%
\url{https://doi.org/10.1038/srep39237}
\showURL{%
\tempurl}


\bibitem[Li et~al\mbox{.}(2017)]%
        {li2017scored}
\bibfield{author}{\bibinfo{person}{Taibo Li}, \bibinfo{person}{Rasmus
  Wernersson}, \bibinfo{person}{Rasmus~B Hansen}, \bibinfo{person}{Heiko Horn},
  \bibinfo{person}{Johnathan Mercer}, \bibinfo{person}{Greg Slodkowicz},
  \bibinfo{person}{Christopher~T Workman}, \bibinfo{person}{Olga Rigina},
  \bibinfo{person}{Kristoffer Rapacki}, \bibinfo{person}{Hans~H St{\ae}rfeldt},
  {et~al\mbox{.}}} \bibinfo{year}{2017}\natexlab{}.
\newblock \showarticletitle{A Scored Human Protein--Protein Interaction Network
  to Catalyze Genomic Interpretation}.
\newblock \bibinfo{journal}{\emph{Nature methods}} \bibinfo{volume}{14},
  \bibinfo{number}{1} (\bibinfo{year}{2017}), \bibinfo{pages}{61--64}.
\newblock
\urldef\tempurl%
\url{https://doi.org/10.1038/nmeth.4083}
\showURL{%
\tempurl}


\bibitem[Li et~al\mbox{.}(2016)]%
        {Li2016GCN}
\bibfield{author}{\bibinfo{person}{Yujia Li}, \bibinfo{person}{Daniel Tarlow},
  \bibinfo{person}{Marc Brockschmidt}, {and} \bibinfo{person}{Richard~S.
  Zemel}.} \bibinfo{year}{2016}\natexlab{}.
\newblock \showarticletitle{Gated Graph Sequence Neural Networks}. In
  \bibinfo{booktitle}{\emph{4th International Conference on Learning
  Representations, {ICLR} 2016}}. \bibinfo{publisher}{ICLR},
  \bibinfo{address}{San Juan, Puerto Rico}.
\newblock
\urldef\tempurl%
\url{http://arxiv.org/abs/1511.05493}
\showURL{%
\tempurl}


\bibitem[Lin(2004)]%
        {lin2004rouge}
\bibfield{author}{\bibinfo{person}{Chin-Yew Lin}.}
  \bibinfo{year}{2004}\natexlab{}.
\newblock \showarticletitle{{ROUGE}: A Package for Automatic Evaluation of
  Summaries}. In \bibinfo{booktitle}{\emph{Text Summarization Branches Out}}.
  \bibinfo{publisher}{Association for Computational Linguistics},
  \bibinfo{address}{Barcelona, Spain}, \bibinfo{pages}{74--81}.
\newblock
\urldef\tempurl%
\url{https://aclanthology.org/W04-1013}
\showURL{%
\tempurl}


\bibitem[Lin et~al\mbox{.}(2015)]%
        {Lin2015HRNN}
\bibfield{author}{\bibinfo{person}{Rui Lin}, \bibinfo{person}{Shujie Liu},
  \bibinfo{person}{Muyun Yang}, \bibinfo{person}{Mu Li}, \bibinfo{person}{Ming
  Zhou}, {and} \bibinfo{person}{Sheng Li}.} \bibinfo{year}{2015}\natexlab{}.
\newblock \showarticletitle{Hierarchical Recurrent Neural Network for Document
  Modeling}. In \bibinfo{booktitle}{\emph{Proceedings of the 2015 Conference on
  Empirical Methods in Natural Language Processing, {EMNLP} 2015}}.
  \bibinfo{publisher}{The Association for Computational Linguistics},
  \bibinfo{address}{Lisbon, Portugal}, \bibinfo{pages}{899--907}.
\newblock
\urldef\tempurl%
\url{https://doi.org/10.18653/v1/d15-1106}
\showDOI{\tempurl}


\bibitem[Liu et~al\mbox{.}(2019)]%
        {Liu2019Paraphrase}
\bibfield{author}{\bibinfo{person}{Yuanxin Liu}, \bibinfo{person}{Zheng Lin},
  \bibinfo{person}{Fenglin Liu}, \bibinfo{person}{Qinyun Dai}, {and}
  \bibinfo{person}{Weiping Wang}.} \bibinfo{year}{2019}\natexlab{}.
\newblock \showarticletitle{Generating Paraphrase with Topic as Prior
  Knowledge}. In \bibinfo{booktitle}{\emph{Proceedings of the 28th {ACM}
  International Conference on Information and Knowledge Management, {CIKM}
  2019}}. \bibinfo{publisher}{{ACM}}, \bibinfo{address}{Beijing, China},
  \bibinfo{pages}{2381--2384}.
\newblock
\urldef\tempurl%
\url{https://doi.org/10.1145/3357384.3358102}
\showDOI{\tempurl}


\bibitem[Liu et~al\mbox{.}(2021)]%
        {liu2021graphine}
\bibfield{author}{\bibinfo{person}{Zequn Liu}, \bibinfo{person}{Shukai Wang},
  \bibinfo{person}{Yiyang Gu}, \bibinfo{person}{Ruiyi Zhang},
  \bibinfo{person}{Ming Zhang}, {and} \bibinfo{person}{Sheng Wang}.}
  \bibinfo{year}{2021}\natexlab{}.
\newblock \showarticletitle{Graphine: A Dataset for Graph-aware Terminology
  Definition Generation}.
\newblock \bibinfo{journal}{\emph{arXiv preprint arXiv:2109.04018}}
  (\bibinfo{year}{2021}).
\newblock


\bibitem[Luong et~al\mbox{.}(2015)]%
        {Luong2015Seq2Seq_Attention}
\bibfield{author}{\bibinfo{person}{Thang Luong}, \bibinfo{person}{Hieu Pham},
  {and} \bibinfo{person}{Christopher~D. Manning}.}
  \bibinfo{year}{2015}\natexlab{}.
\newblock \showarticletitle{Effective Approaches to Attention-based Neural
  Machine Translation}. In \bibinfo{booktitle}{\emph{Proceedings of the 2015
  Conference on Empirical Methods in Natural Language Processing, {EMNLP}
  2015}}. \bibinfo{publisher}{The Association for Computational Linguistics},
  \bibinfo{address}{Lisbon, Portugal}, \bibinfo{pages}{1412--1421}.
\newblock
\urldef\tempurl%
\url{https://doi.org/10.18653/v1/d15-1166}
\showDOI{\tempurl}


\bibitem[Mazandu et~al\mbox{.}(2017)]%
        {Mazandu2017Gene}
\bibfield{author}{\bibinfo{person}{Gaston~K. Mazandu},
  \bibinfo{person}{Emile~R. Chimusa}, {and} \bibinfo{person}{Nicola~J.
  Mulder}.} \bibinfo{year}{2017}\natexlab{}.
\newblock \showarticletitle{Gene Ontology Semantic Similarity Tools: Survey on
  Features and Challenges for Biological Knowledge Discovery}.
\newblock \bibinfo{journal}{\emph{Briefings in Bioinformatics}}
  \bibinfo{volume}{18}, \bibinfo{number}{5} (\bibinfo{year}{2017}),
  \bibinfo{pages}{886--901}.
\newblock
\urldef\tempurl%
\url{https://doi.org/10.1093/bib/bbw067}
\showDOI{\tempurl}


\bibitem[Mutowo et~al\mbox{.}(2016)]%
        {mutowo2016drug}
\bibfield{author}{\bibinfo{person}{Prudence Mutowo},
  \bibinfo{person}{A~Patr{\'\i}cia Bento}, \bibinfo{person}{Nathan Dedman},
  \bibinfo{person}{Anna Gaulton}, \bibinfo{person}{Anne Hersey},
  \bibinfo{person}{Jane Lomax}, {and} \bibinfo{person}{John~P Overington}.}
  \bibinfo{year}{2016}\natexlab{}.
\newblock \showarticletitle{A Drug Target Slim: Using Gene Ontology and Gene
  Ontology Annotations to Navigate Protein-ligand Target Space in ChEMBL}.
\newblock \bibinfo{journal}{\emph{Journal of biomedical semantics}}
  \bibinfo{volume}{7} (\bibinfo{year}{2016}), \bibinfo{pages}{59}.
\newblock
\urldef\tempurl%
\url{https://doi.org/10.1186/s13326-016-0102-0}
\showDOI{\tempurl}


\bibitem[{\"O}zt{\"u}rk et~al\mbox{.}(2020)]%
        {ozturk2020exploring}
\bibfield{author}{\bibinfo{person}{Hakime {\"O}zt{\"u}rk},
  \bibinfo{person}{Arzucan {\"O}zg{\"u}r}, \bibinfo{person}{Philippe
  Schwaller}, \bibinfo{person}{Teodoro Laino}, {and} \bibinfo{person}{Elif
  Ozkirimli}.} \bibinfo{year}{2020}\natexlab{}.
\newblock \showarticletitle{Exploring Chemical Space using Natural Language
  Processing Methodologies for Drug Discovery}.
\newblock \bibinfo{journal}{\emph{Drug Discovery Today}} \bibinfo{volume}{25},
  \bibinfo{number}{4} (\bibinfo{year}{2020}), \bibinfo{pages}{689--705}.
\newblock
\showISSN{1359-6446}
\urldef\tempurl%
\url{https://doi.org/10.1016/j.drudis.2020.01.020}
\showDOI{\tempurl}


\bibitem[Papineni et~al\mbox{.}(2002)]%
        {papineni2002bleu}
\bibfield{author}{\bibinfo{person}{Kishore Papineni}, \bibinfo{person}{Salim
  Roukos}, \bibinfo{person}{Todd Ward}, {and} \bibinfo{person}{Wei{-}Jing
  Zhu}.} \bibinfo{year}{2002}\natexlab{}.
\newblock \showarticletitle{{BLEU}: a Method for Automatic Evaluation of
  Machine Translation}. In \bibinfo{booktitle}{\emph{Proceedings of the 40th
  Annual Meeting of the Association for Computational Linguistics}}.
  \bibinfo{publisher}{{ACL}}, \bibinfo{address}{Philadelphia, PA, {USA}},
  \bibinfo{pages}{311--318}.
\newblock
\urldef\tempurl%
\url{https://doi.org/10.3115/1073083.1073135}
\showDOI{\tempurl}


\bibitem[Peng et~al\mbox{.}(2016)]%
        {Peng2016Extending}
\bibfield{author}{\bibinfo{person}{Jiajie Peng}, \bibinfo{person}{Tao Wang},
  \bibinfo{person}{Jixuan Wang}, \bibinfo{person}{Yadong Wang}, {and}
  \bibinfo{person}{Jin Chen}.} \bibinfo{year}{2016}\natexlab{}.
\newblock \showarticletitle{Extending Gene Ontology with Gene Association
  Networks}.
\newblock \bibinfo{journal}{\emph{Bioinformatics}} \bibinfo{volume}{32},
  \bibinfo{number}{8} (\bibinfo{year}{2016}), \bibinfo{pages}{1185--1194}.
\newblock
\urldef\tempurl%
\url{https://doi.org/10.1093/bioinformatics/btv712}
\showDOI{\tempurl}


\bibitem[Shahab(2017)]%
        {Elham2017Survey}
\bibfield{author}{\bibinfo{person}{Elham Shahab}.}
  \bibinfo{year}{2017}\natexlab{}.
\newblock \showarticletitle{A Short Survey of Biomedical Relation Extraction
  Techniques}.
\newblock \bibinfo{journal}{\emph{CoRR}}  \bibinfo{volume}{abs/1707.05850}
  (\bibinfo{year}{2017}).
\newblock
\showeprint[arXiv]{1707.05850}
\urldef\tempurl%
\url{http://arxiv.org/abs/1707.05850}
\showURL{%
\tempurl}


\bibitem[Srivastava et~al\mbox{.}(2014)]%
        {srivastava2014dropout}
\bibfield{author}{\bibinfo{person}{Nitish Srivastava},
  \bibinfo{person}{Geoffrey~E. Hinton}, \bibinfo{person}{Alex Krizhevsky},
  \bibinfo{person}{Ilya Sutskever}, {and} \bibinfo{person}{Ruslan
  Salakhutdinov}.} \bibinfo{year}{2014}\natexlab{}.
\newblock \showarticletitle{Dropout: a simple way to prevent neural networks
  from overfitting}.
\newblock \bibinfo{journal}{\emph{Journal of Machine Learning Research}}
  \bibinfo{volume}{15}, \bibinfo{number}{1} (\bibinfo{year}{2014}),
  \bibinfo{pages}{1929--1958}.
\newblock
\urldef\tempurl%
\url{http://dl.acm.org/citation.cfm?id=2670313}
\showURL{%
\tempurl}


\bibitem[Stelzer et~al\mbox{.}(2016)]%
        {stelzer2016genecards}
\bibfield{author}{\bibinfo{person}{Gil Stelzer}, \bibinfo{person}{Naomi Rosen},
  \bibinfo{person}{Inbar Plaschkes}, \bibinfo{person}{Shahar Zimmerman},
  \bibinfo{person}{Michal Twik}, \bibinfo{person}{Simon Fishilevich},
  \bibinfo{person}{Tsippi~Iny Stein}, \bibinfo{person}{Ron Nudel},
  \bibinfo{person}{Iris Lieder}, \bibinfo{person}{Yaron Mazor},
  {et~al\mbox{.}}} \bibinfo{year}{2016}\natexlab{}.
\newblock \showarticletitle{The GeneCards Suite: From Gene Data Mining to
  Disease Genome Sequence Analyses}.
\newblock \bibinfo{journal}{\emph{Current protocols in bioinformatics}}
  \bibinfo{volume}{54}, \bibinfo{number}{1} (\bibinfo{year}{2016}),
  \bibinfo{pages}{1--30}.
\newblock
\urldef\tempurl%
\url{https://doi.org/10.1002/cpbi.5}
\showURL{%
\tempurl}


\bibitem[{The Gene Ontology Consortium}(2014)]%
        {GO2015GO}
\bibfield{author}{\bibinfo{person}{{The Gene Ontology Consortium}}.}
  \bibinfo{year}{2014}\natexlab{}.
\newblock \showarticletitle{Gene Ontology Consortium: Going Forward}.
\newblock \bibinfo{journal}{\emph{Nucleic Acids Research}}
  \bibinfo{volume}{43}, \bibinfo{number}{D1} (\bibinfo{date}{11}
  \bibinfo{year}{2014}), \bibinfo{pages}{D1049--D1056}.
\newblock
\showISSN{0305-1048}
\urldef\tempurl%
\url{https://doi.org/10.1093/nar/gku1179}
\showDOI{\tempurl}


\bibitem[{The Gene Ontology Consortium}(2017)]%
        {GO2017GO}
\bibfield{author}{\bibinfo{person}{{The Gene Ontology Consortium}}.}
  \bibinfo{year}{2017}\natexlab{}.
\newblock \showarticletitle{Expansion of the Gene Ontology Knowledgebase and
  Resources}.
\newblock \bibinfo{journal}{\emph{Nucleic Acids Research}}
  \bibinfo{volume}{45}, \bibinfo{number}{Database-Issue}
  (\bibinfo{year}{2017}), \bibinfo{pages}{D331--D338}.
\newblock
\urldef\tempurl%
\url{https://doi.org/10.1093/nar/gkw1108}
\showDOI{\tempurl}


\bibitem[{The UniProt Consortium}(2021)]%
        {UniProt}
\bibfield{author}{\bibinfo{person}{{The UniProt Consortium}}.}
  \bibinfo{year}{2021}\natexlab{}.
\newblock \showarticletitle{UniProt: the Universal Protein Knowledgebase in
  2021}.
\newblock \bibinfo{journal}{\emph{Nucleic Acids Res.}} \bibinfo{volume}{49},
  \bibinfo{number}{Database-Issue} (\bibinfo{year}{2021}),
  \bibinfo{pages}{D480--D489}.
\newblock
\urldef\tempurl%
\url{https://doi.org/10.1093/nar/gkaa1100}
\showDOI{\tempurl}


\bibitem[Tomczak et~al\mbox{.}(2018)]%
        {tomczak2018interpretation}
\bibfield{author}{\bibinfo{person}{Aurelie Tomczak},
  \bibinfo{person}{Jonathan~M Mortensen}, \bibinfo{person}{Rainer Winnenburg},
  \bibinfo{person}{Charles Liu}, \bibinfo{person}{Dominique~T Alessi},
  \bibinfo{person}{Varsha Swamy}, \bibinfo{person}{Francesco Vallania},
  \bibinfo{person}{Shane Lofgren}, \bibinfo{person}{Winston Haynes},
  \bibinfo{person}{Nigam~H Shah}, {et~al\mbox{.}}}
  \bibinfo{year}{2018}\natexlab{}.
\newblock \showarticletitle{Interpretation of Biological Experiments Changes
  with Evolution of the Gene Ontology and its Annotations}.
\newblock \bibinfo{journal}{\emph{Scientific reports}} \bibinfo{volume}{8},
  \bibinfo{number}{1} (\bibinfo{year}{2018}), \bibinfo{pages}{1--10}.
\newblock
\urldef\tempurl%
\url{https://doi.org/10.1038/s41598-018-23395-2}
\showURL{%
\tempurl}


\bibitem[Vaswani et~al\mbox{.}(2017)]%
        {Vaswani2017Transformer}
\bibfield{author}{\bibinfo{person}{Ashish Vaswani}, \bibinfo{person}{Noam
  Shazeer}, \bibinfo{person}{Niki Parmar}, \bibinfo{person}{Jakob Uszkoreit},
  \bibinfo{person}{Llion Jones}, \bibinfo{person}{Aidan~N. Gomez},
  \bibinfo{person}{Lukasz Kaiser}, {and} \bibinfo{person}{Illia Polosukhin}.}
  \bibinfo{year}{2017}\natexlab{}.
\newblock \showarticletitle{Attention is All you Need}. In
  \bibinfo{booktitle}{\emph{Advances in Neural Information Processing Systems
  30: Annual Conference on Neural Information Processing Systems 2017, December
  4-9, 2017}}. \bibinfo{publisher}{Curran Associates, Inc.},
  \bibinfo{address}{Long Beach, CA, {USA}}, \bibinfo{pages}{5998--6008}.
\newblock
\urldef\tempurl%
\url{https://proceedings.neurips.cc/paper/2017/hash/3f5ee243547dee91fbd053c1c4a845aa-Abstract.html}
\showURL{%
\tempurl}


\bibitem[Wang et~al\mbox{.}(2022)]%
        {wang2022textomics}
\bibfield{author}{\bibinfo{person}{Mu-Chun Wang}, \bibinfo{person}{Zixuan Liu},
  {and} \bibinfo{person}{Sheng Wang}.} \bibinfo{year}{2022}\natexlab{}.
\newblock \showarticletitle{Textomics: A dataset for genomics data summary
  generation}. In \bibinfo{booktitle}{\emph{Proceedings of the 60th Annual
  Meeting of the Association for Computational Linguistics (Volume 1: Long
  Papers)}}. \bibinfo{pages}{4878--4891}.
\newblock


\bibitem[Wang et~al\mbox{.}(2015)]%
        {wang2015exploiting}
\bibfield{author}{\bibinfo{person}{Sheng Wang}, \bibinfo{person}{Hyunghoon
  Cho}, \bibinfo{person}{ChengXiang Zhai}, \bibinfo{person}{Bonnie Berger},
  {and} \bibinfo{person}{Jian Peng}.} \bibinfo{year}{2015}\natexlab{}.
\newblock \showarticletitle{Exploiting Ontology Graph for Predicting Sparsely
  Annotated Gene Function}.
\newblock \bibinfo{journal}{\emph{Bioinformatics}} \bibinfo{volume}{31},
  \bibinfo{number}{12} (\bibinfo{year}{2015}), \bibinfo{pages}{357--364}.
\newblock
\urldef\tempurl%
\url{https://doi.org/10.1093/bioinformatics/btv260}
\showDOI{\tempurl}


\bibitem[Wang et~al\mbox{.}(2021)]%
        {wang2021leveraging}
\bibfield{author}{\bibinfo{person}{Sheng Wang},
  \bibinfo{person}{Angela~Oliveira Pisco}, \bibinfo{person}{Aaron McGeever},
  \bibinfo{person}{Maria Brbic}, \bibinfo{person}{Marinka Zitnik},
  \bibinfo{person}{Spyros Darmanis}, \bibinfo{person}{Jure Leskovec},
  \bibinfo{person}{Jim Karkanias}, {and} \bibinfo{person}{Russ~B Altman}.}
  \bibinfo{year}{2021}\natexlab{}.
\newblock \showarticletitle{Leveraging the Cell Ontology to classify unseen
  cell types}.
\newblock \bibinfo{journal}{\emph{Nature communications}} \bibinfo{volume}{12},
  \bibinfo{number}{1} (\bibinfo{year}{2021}), \bibinfo{pages}{1--11}.
\newblock


\bibitem[Xu and Wang(2022)]%
        {xu2022protranslator}
\bibfield{author}{\bibinfo{person}{Hanwen Xu} {and} \bibinfo{person}{Sheng
  Wang}.} \bibinfo{year}{2022}\natexlab{}.
\newblock \showarticletitle{ProTranslator: zero-shot protein function
  prediction using textual description}. In
  \bibinfo{booktitle}{\emph{International Conference on Research in
  Computational Molecular Biology}}. Springer, \bibinfo{pages}{279--294}.
\newblock


\bibitem[Yang et~al\mbox{.}(2022)]%
        {yangpathway2text}
\bibfield{author}{\bibinfo{person}{Junwei Yang}, \bibinfo{person}{Zequn Liu},
  \bibinfo{person}{Ming Zhang}, {and} \bibinfo{person}{Sheng Wang}.}
  \bibinfo{year}{2022}\natexlab{}.
\newblock \showarticletitle{Pathway2Text: Dataset and Method for Biomedical
  Pathway Description Generation}.
\newblock  (\bibinfo{year}{2022}).
\newblock


\bibitem[Zhang et~al\mbox{.}(2020)]%
        {Zhang2020GO}
\bibfield{author}{\bibinfo{person}{Yanjian Zhang}, \bibinfo{person}{Qin Chen},
  \bibinfo{person}{Yiteng Zhang}, \bibinfo{person}{Zhongyu Wei},
  \bibinfo{person}{Yixu Gao}, \bibinfo{person}{Jiajie Peng},
  \bibinfo{person}{Zengfeng Huang}, \bibinfo{person}{Weijian Sun}, {and}
  \bibinfo{person}{Xuanjing Huang}.} \bibinfo{year}{2020}\natexlab{}.
\newblock \showarticletitle{Automatic Term Name Generation for Gene Ontology:
  Task and Dataset}. In \bibinfo{booktitle}{\emph{Findings of the Association
  for Computational Linguistics: {EMNLP} 2020}}.
  \bibinfo{publisher}{Association for Computational Linguistics},
  \bibinfo{address}{Online}, \bibinfo{pages}{4705--4710}.
\newblock
\urldef\tempurl%
\url{https://doi.org/10.18653/v1/2020.findings-emnlp.422}
\showDOI{\tempurl}


\end{thebibliography}

\end{document}